# ENHANCING BORDER SECURITY AND COUNTERING TERRORISM THROUGH COMPUTER VISION: A FIELD OF ARTIFICIAL INTELLIGENCE

Tosin Ige[1], Abosede Kolade[2] Olukunle Kolade[3]

[1] University of Texas at El Paso, Texas, TX 79968, USA
[2] Texas A&M University-Commerce, Texas, TX, 75428, USA
[3] United State Navy, 1200 Navy, Pentagon, Washington, DC, 20350-1200, USA

**Abstract.** Border security had been a persistent problem in international border especially when it get to the issue of preventing illegal movement of weapons, contraband, drugs, and combating issue of illegal or undocumented immigrant while at the same time ensuring that lawful trade, economic prosperity coupled with national sovereignty across the border is maintained.

In this research work, we used open source computer vision (Open CV) and adaboost algorithm to develop a model which can detect a moving object a far off, classify it, automatically snap full image and face of the individual separately, and then run a background check on them against worldwide databases while making a prediction about an individual being a potential threat, intending immigrant, potential terrorists or extremist and then raise sound alarm. Our model can be deployed on any camera device and be mounted at any international border.

There are two stages involved, we first developed a model based on open CV computer vision algorithm, with the ability to detect human movement from afar, it will automatically snap both the face and the full image of the person separately, and the second stage is the automatic triggering of background check against the moving object. This ensures it check the moving object against several databases worldwide and is able to determine the admissibility of the person afar off. If the individual is inadmissible, it will automatically alert the border officials with the image of the person and other details, and if the bypass the border officials, the system is able to detect and alert the authority with his images and other details. All these operations will be done afar off by the AI powered camera before the individual reach the border.

## 1 Introduction

Recent advancement in artificial intelligence had led to computer vision as a subfield of artificial intelligence thereby enabling computer to derive and analysis information from visual data (images, videos and several other graphical) inputs, these advancements had position Computer vision which is a field of artificial intelligence as ultimate solution to border insecurity, maritime insecurity, terrorism, airport arrival and departure point.

Insecurity is the condition in which a nation cannot adequately defend herself against threats arising from danger or aggression which can jeopardize her independence and or territory. Insecurity has no border and it had grown to being a global threat which consistently threatened our peaceful co-existence and international relation. There are several causes of international insecurity across border, some of which includes but not limited to balancing power between superpowers, pursuing of arms race, constant conflict, territorial claim and terrorism.

The incidence of September 11, 2021 multiple terrorist attacks on the United States leads to formation of the department of homeland security, whose mission is saddled with counter terrorism, cyber security, border security, and also natural disaster planning and response (Fig. 1)[13]. With the sum of $133 billion having been spent since the creation of department of homeland security in 2003 by the US government, border security and enforcement remains a big challenge despite the increase in the number of border security and interior enforcement officer exceeding 50,000. In the same vein, The EU faces a number of threats at the external borders, stemming from factors such as irregular migration, people smuggling, human trafficking, illegal weapons, and drugs [1]. Artificial Intelligence (AI) yields great potential to significantly enhance the effectiveness of security and operational activities at the borders and beyond, with direct and multiplying impact on internal security.

It was these challenges that led to the commission of RAND Europe by the Frontex [6],[7] to explore and search for ways by which European borders can adopt the use of artificial intelligence to secure international borders within the European union [2], while at the same time looking at the prospects of developing and deploying fair and secure AI with capability to maximize the support for the management of European Union's border.
.

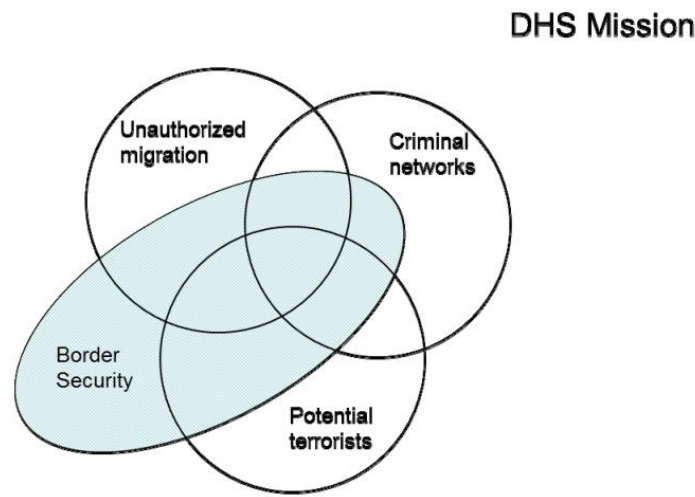

Fig 1. Border threats and DHS Mission

Despite the fact that billions of Dollars is being budgeted annually for department of homeland security to counter terrorism, secure border, defense cyber security, respond to natural disaster, cracking down on criminal networks, identification and prevention of unauthorized migration and potential terrorists. The problem remains daunting and unsolved yet by the department of homeland security as evidence by recent statistic from Pen research center (Fig 2)[14];

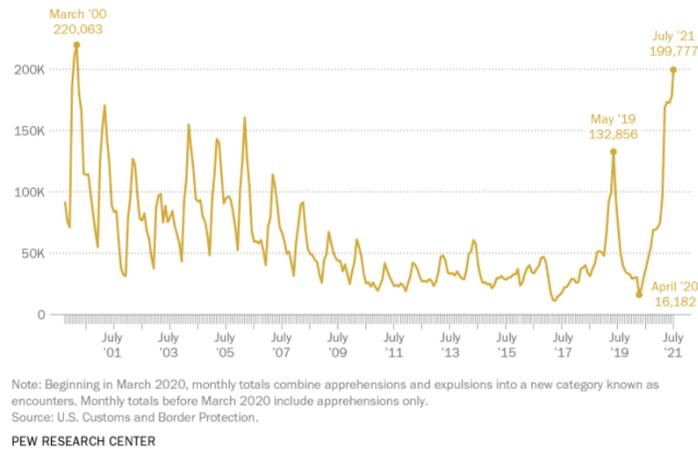

Fig 2. Migrant encounters at U.S.-Mexico border are at a 21-year high

In the same vein, terrorism on its own which involves the use of threats of violence by non state actor to achieve their goals which may be political, economical, or to force international relationship. Data from the Center for Specific and International studies shows annual geometric increment in the percentage of terrorists' attacks and plots in the United States between 2015 and 2020 (Fig. 3)[16].

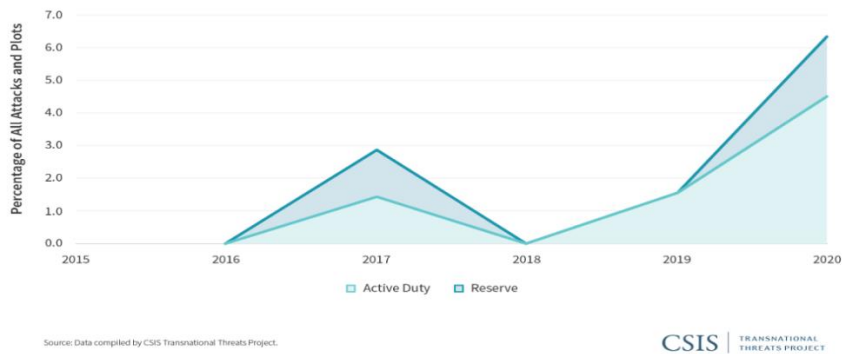

Fig 3. Percentage of U.S. Terrorist attacks and plots perpetrated by Active Duty, or Reserve Service Members, 2015-2020

Currently at various international borders, AI is only being used as complements for border officials to monitor and scan migrants and territory. This current usage of AI at securing border is not effective considering the possibility of evasion at the border by illegal migrants, also there is the question of how to control influx of terrorist from a very large border of several kilometers long, and this increases the porosity. In order to effectively control and man an international border, there is need to automate an AI powered control tower decorated with that is able to detect movement afar off along with automated background check before the migrant or terrorist reach the border.

Problem of securing border, counter terrorism, and illegal immigrant migration becomes more challenging and daunting due to frequently changing tactics adopted by potential immigrants to bit security system at airport and land border. It is for this purpose that we developed a machine learning model using artificial neural network and adaBoost algorithm for image identifier and object recognition. Two algorithms are implemented in the models which are neural network and adaBoost. The adaBoost algorithm implementation enables the recognition and identification of walking objects across the border while the artificial neural network implementation scan deep and analysis each and every objects in the body of the image afar off and then make informed prediction.

## 2    Background Study

Current background study on border security can me categorized into two; the first category is of the opinion that it is inevitable to impose security on peoples movement so far the likelihood of illegal, criminal, and terrorism activities is at the barest minimum, the other category is of the opinion that subjecting people at the border crossing to extensive security control not a necessity rather than tends to subjects people to unnecessary stress and human right violation [3].
This category represents mostly governmental approaches that can sometimes be based on public opinion such as the euro barometer survey [4] and a host of other reasons to secure their international border [5].

It is also the fundamental duty of the government to secure its border to minimize terrorist activities, illegal migration, and human and drug trafficking, etc. While several artificial intelligence had been deployed at various borders and airports, there had been very little or no success as evidence by the series of illegal migrant at various international border, terrorism, and extremism. Harel [8] did a good job in raising ethical issues in deploying artificial intelligence but his concern is based on biometric identification, while Hayes  and Vermeulen [9] were only concerned with the economic cost and fundamental implication it will have on human right while failing to proffer solution to the problem.

The fact that some artificial intelligence had been implemented at various international land border [10] for border control with little success cannot be over emphasize due its ability to infringe and tamper with the fundamental human right[11],[12] of people at border crossing and checkpoint.  In addition to the fact that implemented AI at some of the international border has little success, it constantly infringe on people human fundamental human righ like the lie detector, mine reader, etc. This lead to our development of a new computer vision AI which is based on Open source computer vision algorithm without any infringement on privacy

## 3    Research Methodology

 For this research work we use;
1.  Open source computer vision (OpenCV) Library
2.  Face Recognition library
3.  Live stream video from Webcam
4.  Python 3.10.2
5.  MSSQL Server Database
6.  Restful API service

Due to the massive movements in illegal immigrant, terrorism, and criminality against humanity which had resulted in insecurity across the globe. There is need for an artificial intelligence powered solution that will detect, track and raise ring by notifying appropriate authority around on any potential terrorist, extremist, illegal immigrant, drug cartel, or someone on the wanted list of Interpol or FBI.
In our approach to implementing artificial intelligence that is based on computer vision to proffers permanent solution to this, we wrote a machine learning algorithm that is able to detect image in a live streaming camera using open source computer vision machine learning algorithm. We wrote two separate algorithms namely;
i.   Human-Detection algorithm
ii.  Face-Extraction algorithm

The function of the human detection algorithm is to detect human in a live stream camera and then save the image in a particular folder on the system. The face-extraction algorithm role is to complement the work of the human-detection algorithm by constantly checking the system for any saved images, and then extract the face part into another location.

We developed a standalone restful API service using python FLASK framework, the restful web service takes the faces, convert them individually to bite array and then use the converted bite array to make a thorough background search against several and security agents' database such as police database, public database, government database, immigration database, Interpol, etc. If the person is on wanted list, the service is able to notify the appropriate officers at the border while the individual is still afar off, should incase the individual find his way and beat the security officer at the border, the system is able to detect, log it on the database, notify security authority, and also raise necessary alarm. Since the program is artificial intelligence based using computer vision algorithm. All that is needed is to mount an external camera or webcam, our system is able to detect human being in the camera and save it immediately. If it is a group of people that are present in the camera surveillance, it will split them into individual group, extract their faces individually, and convert each faces to separate byte array, before saving it to database. As soon as it hit the database, another restful API service is triggered which pick the byte array and then make a thorough search against several database for background check.

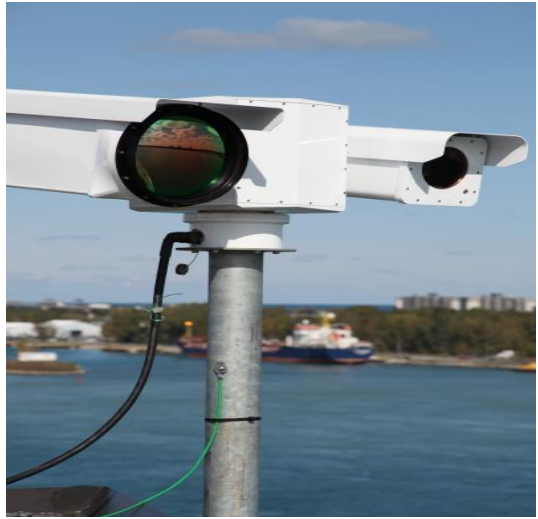

Fig 4. Video surveillance system on Port Huron Michigan Photo by Health Stephens

The flow chart is such that once the surveillance camera (Fig 4)[12] detects an image, the system automatically classified it into, object, animal, and human, if the image is categorized as object or animal, the system ignores, but if it is human, it will extract the face of the person, while automatically triggering a specially built restful application programming interface (API) algorithm that converts the face to a bite array and then do a thorough background check from multiple worldwide database such as social media ( for terrorism or extremism view), immigration (Document of undocumented) record, police record (criminal record), inter-pol (wanted list) using asynchronous method. All process happen in couple of seconds and then the system is able to classify the result into two categories (admissible or inadmissible). If the result of all those background check against multiple worldwide database is good, then it is classify as admissible which will be ignore by the system, but if the result is otherwise not good, the image is classify as inadmissible, and so the system automatically report through a dedication channel, send alert and sound alarm to the officials at the border.

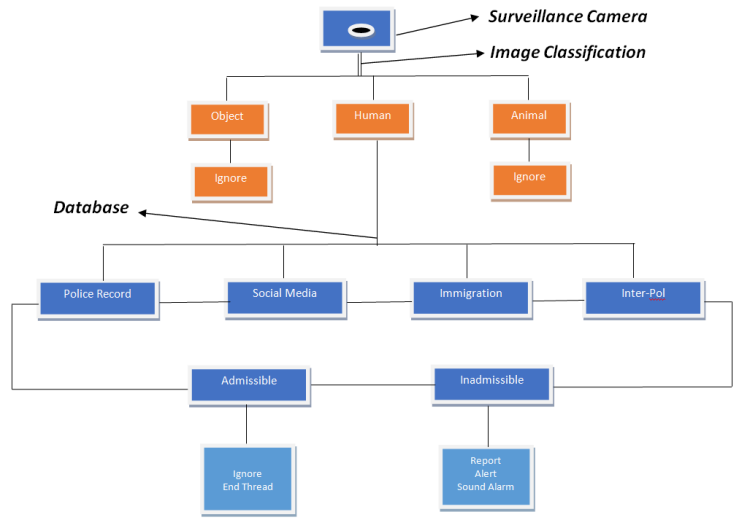

Fig 5. Border Security Application flow chart Diagram

The whole process is based on multi-threading as there can be hundreds of people crossing the border at the same time, hence, the system uses asynchronous method to run multiple tasks underground at the same time against each image (Fig 5).

```python
import cv2

# Load the cascade
face_cascade = cv2.CascadeClassifier('haarcascade_frontalface_default.xml')

# To capture video from webcam.
cap = cv2.VideoCapture(0)

while True:
    # Read the frame
    _, img = cap.read()
    # Convert to grayscale
    gray = cv2.cvtColor(img, cv2.COLOR_BGR2GRAY)
    # Detect the faces
    faces = face_cascade.detectMultiScale(gray, 1.1, 4)
    # Draw the rectangle around each face
    for (x, y, w, h) in faces:
        cv2.rectangle(img, (x, y), (x+w, y+h), (255, 0, 0), 2)
    # Display
    cv2.imshow('img', img)
    # Filename
    filename = 'savedImage.jpg'

    # Using cv2.imwrite() method
    # Saving the image
    cv2.imwrite(filename, img)

    # Stop if escape key is pressed
    k = cv2.waitKey(30) & 0xff
    if k==27:
        break
# Release the VideoCapture object
cap.release()
out.release()
cv2.destroyAllWindows()
```

Fig 6. Algorithm for image detection and face extraction from Surveillance camera

The program is based on Open Source Computer Vision (Open CV) algorithm (Fig 6.), with the capability to detect, capture and extract face from image in a live stream or surveillance camera.

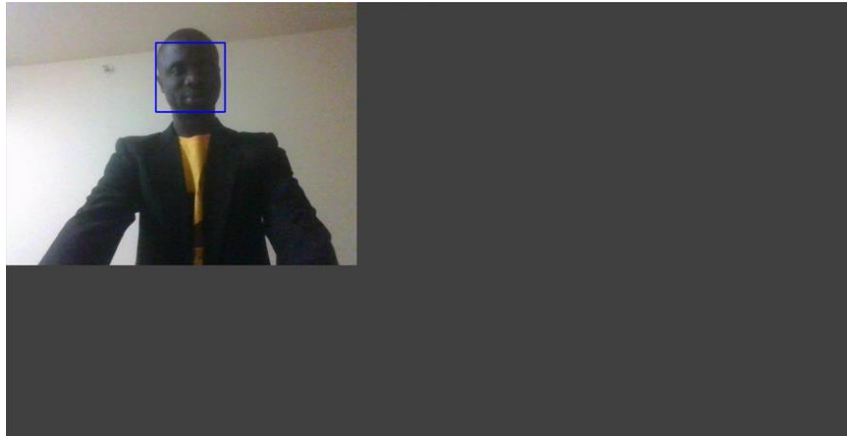

Fig 7. Real Time Face Detection and Extraction from Live surveillance Camera

On successful detection and extraction of face from image in the camera (Fig 7), another background service in the system pick the extracted faces, convert it to byte array, and check against various worldwide database to determine the admissibility or inadmissibility of the person to the country.

### Performance Evaluation and Validation

For our validation and performance evaluation, we use several images at different distances to look for possible false positive and false negative. We also use group of people walking side by side and also separately for test. In all we have true positive and true negative in each case which signifies the presence of low variance and low bias in our model. The only area where we have false positive is when the person is wearing a cloth which has image of a person on it, the system pick both the face of the person as well as the face of the image printed on the cloth.

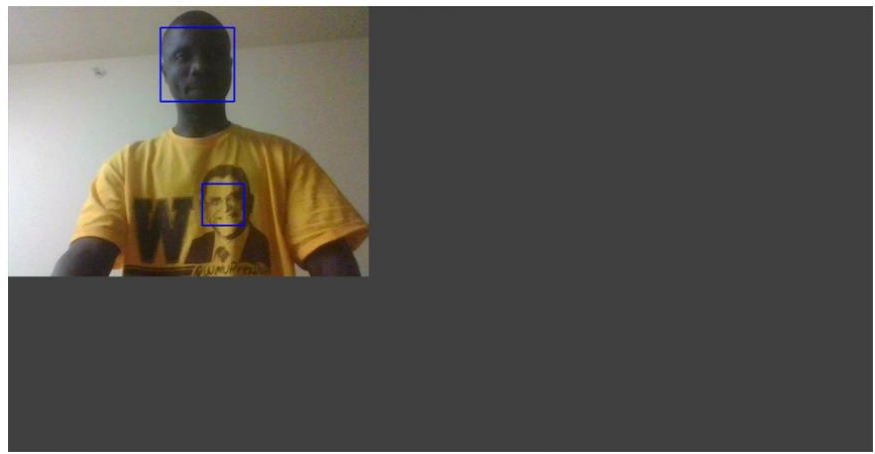

Fig 8. True Positive and False Positive from Real Time Face Detection and Extraction from Live surveillance Camera

So, here we have both true positive and false positive, extracting the face of the image printed on the cloth is false positive, this is going to stress the system to some extent because the restful API services algorithm will do background check on it against various databases. It is of my opinion that a new model can be developed in the near future to automatically detect and exclude images printed on the cloth so as to eliminate the false positive scenario.

## Conclusion

In this research work, we used open source computer vision (Open CV) algorithm to develop a model which can detect a moving object a far off, classify it into animate object, inanimate object or human, automatically snap full image and face of the individual separately in the process, and then run a background check on them against worldwide databases and then make a prediction about an individual being a potential threat, intending immigrant, potential terrorists or extremist and then sound alarm. Our model can be deployed on any camera device and be mounted at any international border. With this research and implementation which is based on computer vision branch of artificial intelligence, the problems of border security which had been a persistent problem in international border especially when it get to the issue of preventing illegal movement of weapons, contraband, drugs, and combating issue of illegal or undocumented immigrant can be effectively solve while at the same time ensuring that lawful trade, economic prosperity coupled with national sovereignty across the border is maintained and enhanced.

Since, this research is based on computer vision, we used use true positive and false positive for our (Fig 8) performance evaluation and cross validation which was generally good and deployable. The only area where we have false positive is if the person is wearing a shirt or cloth with an image on it, the system will extract both the face of the person as well as the face of the image printed on the shirt. We categorize this as false positive because the system will automatically trigger and run a background check against the extracted face from the printed image on the shirt. So, for someone wearing cloth with image printed on it, the system will perform background check on both the person and the image printed on his cloth. Then, what happen if the person is admissible but the image of the person printed on the shirt is a terrorist and inadmissible? It means there will be false alarm and wrong notification at the border. This is an area where more research work is still needed in the near future. There is no risk or danger here because it is just the image from the person printed on the cloth and the actual person is not at the border, we believe further research is needed here as regards the false alarm arising from extracting face from the image printed on the shirt of someone crossing border.

## Limitation

We had limitation in accessing records like Interpol, Federal Bureau of Investigation (FBI), police record, immigration records, and so on because special permission from the appropriate authority is required to access them due to the privacy and sensitivity of such records.

## Data Availability

We had limitation in accessing records like Interpol, Federal Bureau of Investigation (FBI), police record, immigration records, and so on because special permission from the appropriate authority is required to access them due to the privacy and sensitivity of such records.

16. The Military, Police, and the Rise of Terrorism in the United States | Center for Strategic and International Studies
https://www.csis.org/analysis/military-police-and-rise-terrorism-united-states